\documentclass[letterpaper, 10 pt, conference]{ieeeconf}  

\IEEEoverridecommandlockouts                              

\usepackage{graphics} 
\usepackage{epsfig} 
\usepackage{mathptmx} 
\usepackage{times} 
\usepackage{amsmath} 
\usepackage{amssymb}  
\usepackage{booktabs}
\usepackage{graphicx}
\usepackage{booktabs}
\usepackage{cite}
\usepackage{multirow}
\usepackage{algorithm,xspace}
\usepackage{algpseudocode}
\usepackage[T1]{fontenc}
\usepackage{caption}
\let\labelindent\relax
\usepackage{wrapfig,enumitem}

\newcommand{\ours}{\textit{ADEPT}\xspace}
\newcommand{\absrel}{\(\text{Abs}_{\text{rel}}\)\xspace}

\title{\LARGE \bf
ADEPT: Adaptive Monocular Depth Estimation via Model-In-The-Loop Photometric Tuning with Capture Risk Prediction
}

\author{GiHyun Kim$^{1}$, Jin Mo Yang$^{2}$, and Hyung-Sin Kim$^{2}$
\thanks{$^{1}$AIoT Lab, Seoul National University, Seoul, Korea, kimh060618@gmail.com}
\thanks{$^{2}$Seoul National University, Seoul, Korea, \{yjm3412, hyungkim\}@snu.ac.kr}
}

\begin{document}

\raggedbottom

\maketitle
\thispagestyle{empty}
\pagestyle{empty}

\begin{abstract}

Exposure and gain determine the visual evidence available to monocular depth estimation (MDE), particularly when robot motion creates a tradeoff between light collection and motion blur. 
We present \ours, a framework that adapts these settings using learned feedback from a frozen depth model. During training, ground-truth depth identifies the lowest-error reference within a group of physically evaluated camera settings. A compact predictor learns reference-relative disagreement (RRD) between each candidate’s depth prediction and the reference prediction using regression and ranking losses. At deployment, it scores observed captures using frozen-model features and camera, motion, and illumination context, without ground truth or reference images. A contextual Gaussian-process controller uses recent scores to select subsequent exposure and gain settings under motion and photometric constraints. 
We evaluate offline camera selection with relative and metric Depth Anything v2 and demonstrate online control with the affine-invariant variant on a mobile robot. Across nine equally weighted scene–illumination conditions, \ours reduces mean scale-and-shift-aligned \absrel by 14.6\% relative to built-in auto-exposure and by 10.0\% relative to the best-performing evaluated prior controller.

\end{abstract}

\section{Introduction}
\label{sec:introduction}

Monocular depth estimation (MDE) provides dense geometric cues from a single RGB image, making it attractive for mobile robots constrained by sensor cost, size, or payload~\cite{wofk2019fastdepth}.
Recent depth foundation models further expand the potential of RGB cameras as a source of geometry through improved generalization across diverse scenes~\cite{yang2024depth,piccinelli2024unidepth,depthanything3}. Deploying these models reliably, however, depends on \textbf{how RGB images are acquired} as well as on the model that interprets them.

Camera settings determine the visual evidence available \textit{before} inference begins. Saturation can erase intensity differences, while motion blur and low signal-to-noise ratios can obscure boundaries and texture~\cite{shin2019camera,kim2018exposure,shim2018gradient,zhang2017active}. Learned priors can compensate for degraded observations, but faithful recovery of details lost during capture is not assured~\cite{eilertsen2017hdr}. 
These effects are particularly important for moving robots: a long exposure that reveals a dark scene at rest can blur it during a turn, whereas shortening exposure limits the collected light. These tradeoffs motivate \textbf{adapting acquisition} to preserve evidence useful to the depth model as motion and illumination change.

The central challenge is obtaining feedback that distinguishes camera settings by their usefulness to the deployed depth model, without ground-truth depth online. Classical controllers optimize image statistics~\cite{shin2019camera,kim2018exposure,shim2018gradient,zhang2017active}, while learned approaches incorporate supervision from downstream perception tasks~\cite{lee2024learning,zhang2024efficient,tomasi2021learned}. LENS~\cite{baek2025adaptive} further uses a frozen model's confidence to guide acquisition in quasi-static environments. These approaches establish the value of task-aware camera control, but feedback based on appearance, feature matching, or classification does not directly quantify capture quality for MDE. Depth uncertainty estimators provide relevant signals~\cite{kendall2017uncertainties,poggi2020uncertainty,sudhakar2022uncertainty,hornauer2022gradient,hornauer2025revisiting,mi2022training,landgraf2026critical}, but their usefulness for ranking physical camera settings must also be established. On a moving robot, these rankings must be updated from successive observations as viewpoint and illumination change. 

\begin{figure}[t]
    \centering
    \includegraphics[width=\columnwidth]{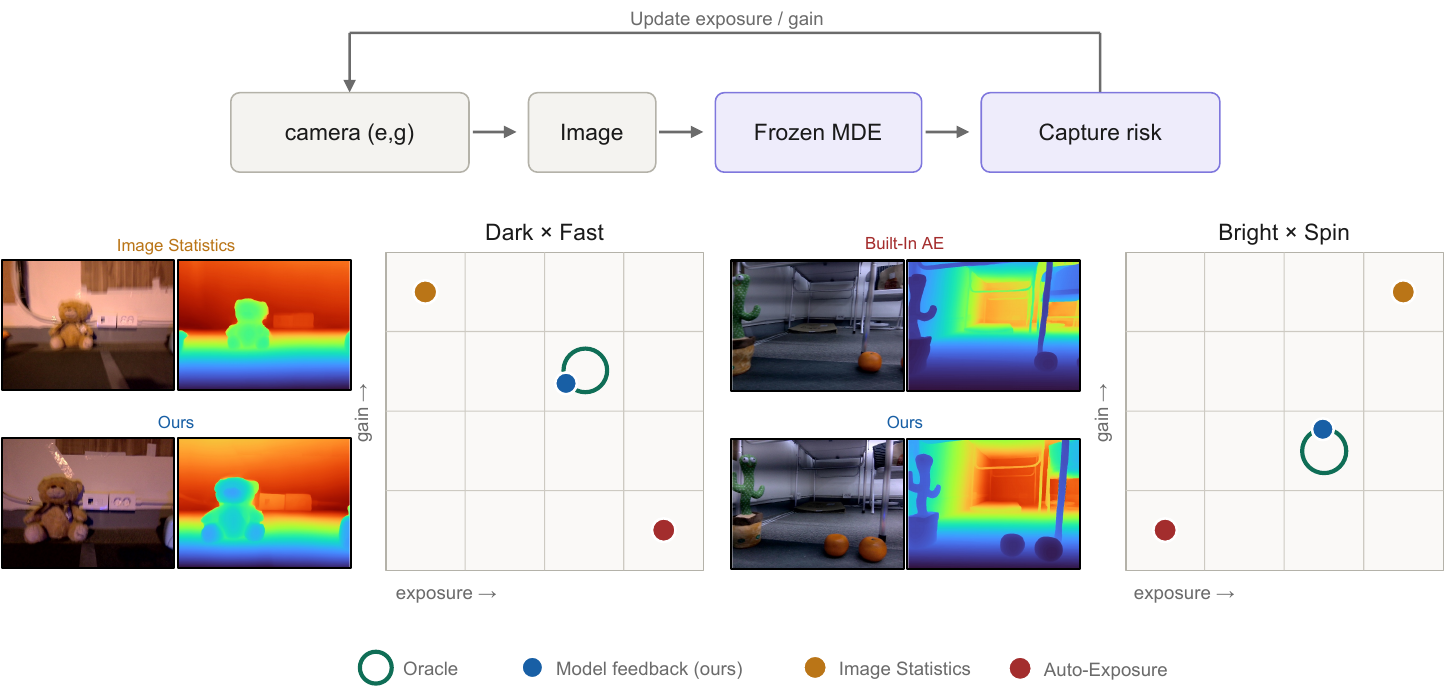}
    \caption{
    Overview of \ours. Learned feedback from acquired images guides exposure and gain updates for a frozen depth model. The lower panels provide illustrative comparisons of selected camera settings under dark$\times$fast and bright$\times$spin conditions with the best evaluated reference.
    }
    \label{fig:intro_framework}
    \vspace{-1mm}
\end{figure}

We introduce \ours, a framework that adapts exposure and gain using \textbf{reference-relative disagreement (RRD)} as learned feedback for a frozen MDE model. 
During training, ground-truth depth identifies the lowest-error reference within a group of captures at approximately matched viewpoints and different camera settings. A compact predictor learns disagreement between each candidate's depth prediction and the reference prediction (RRD), with an auxiliary ranking loss. RRD serves as a surrogate for relative capture quality, whose usefulness is evaluated through camera-selection experiments. 
At deployment, the predictor scores observed captures using frozen MDE features and camera, motion, and illumination context, without ground truth or reference images. A contextual Gaussian-process (GP) controller uses recent scores to select the next exposure and gain under motion-dependent and photometric constraints. 
We evaluate the resulting feedback across training and validation scene configurations and in online mobile-robot experiments.


Our contributions are threefold:
\begin{itemize}[leftmargin=*]
    \item An \textbf{acquisition-oriented supervision target (RRD)}, and a \textbf{compact predictor} that learns to compare camera settings for a frozen MDE model using ground-truth-selected references and a ranking loss.
    \item A \textbf{controller} that integrates this feedback into sequential exposure and gain selection, using contextual Gaussian-process (GP) optimization with motion-dependent and photometric constraints. 
    \item An evaluation of camera-setting selection using relative and metric variants of Depth Anything v2, together with online control on a mobile robot. Across nine scene--illumination conditions, \ours{} reduces mean \absrel by 14.6\% relative to built-in auto-exposure and by 10.0\% relative to Zhang et al.~\cite{zhang2024efficient}, with each condition weighted equally.
\end{itemize}

\section{Related Work}
\label{sec:rel_works}

\subsection{Camera Control for Robotic Perception}

Camera-control methods adapt image acquisition to improve visual perception. Classical approaches optimize hand-crafted image-quality metrics based on gradients, entropy, brightness, or noise statistics~\cite{shin2019camera,kim2018exposure,shim2018gradient,zhang2017active}. Shin et al.~\cite{shin2019camera} combine these measures with Nelder--Mead optimization, while Kim et al.~\cite{kim2018exposure} use Bayesian optimization of an entropy-weighted gradient metric. 
%
Learned approaches incorporate richer supervision: Tomasi et al.~\cite{tomasi2021learned} train exposure and gain control using feature matches from a VO or SLAM pipeline, Lee et al.~\cite{lee2024learning} learn exposure policies with rewards for image attributes and flicker, and Zhang et al.~\cite{zhang2024efficient} use image history and rewards that include visual-odometry performance.

LENS~\cite{baek2025adaptive} uses confidence from a frozen classification model to select among candidate captures in quasi-static environments, with candidate-subset strategies that balance accuracy and acquisition cost. 
In contrast, \ours learns \textit{RRD for MDE camera selection} and uses sequential feedback: its predictor scores the current capture, and a \textit{contextual GP} uses recent scores to select subsequent exposure and gain settings as viewpoint, motion, and illumination change.

\subsection{Uncertainty and Risk Prediction for Monocular Depth Estimation}

MDE uncertainty has been studied through probabilistic prediction, learned estimation, and post-hoc analysis~\cite{sudhakar2022uncertainty,gupta2024estimating,yang2019inferring,yang2020d3vo,mi2022training,hornauer2022gradient,landgraf2026critical,poggi2020uncertainty,kendall2017uncertainties,hornauer2025revisiting}. SLURP~\cite{yu2021slurp} uses a side learner to estimate regression uncertainty from the main model's outputs and intermediate representations. Gradient-based methods~\cite{hornauer2022gradient, hornauer2025revisiting} and test-time augmentation~\cite{mi2022training} offer training-free alternatives.  
\ours's contribution is to supervise an image-level predictor using disagreement with a ground-truth-selected reference capture at an approximately matched viewpoint. Its output serves as a camera-selection cost. We evaluate this acquisition-oriented use through groupwise selection regret and online depth accuracy, with a supervision ablation examining the choice of training target.

\subsection{Monocular Depth for Robotics}

FastDepth~\cite{wofk2019fastdepth} studies embedded depth inference, while D3VO~\cite{yang2020d3vo} integrates learned depth, pose, and uncertainty into visual odometry. Large-scale training improves relative-depth
generalization~\cite{yang2024depth, birkl2023midas,yang2024depth1,Ranftl2022}, and metric-depth models broaden the applicability of single-image geometry~\cite{piccinelli2024unidepth, bhat2023zoedepth, bochkovskiy2025depth, depthanything3}. 
\ours{} complements these model advances by \textit{adapting image acquisition} (exposure and gain) for frozen depth models. 

\section{Empirical Motivation}
\label{sec:observations}

\begin{figure*}[t]
    \centering
    \includegraphics[width=\textwidth]{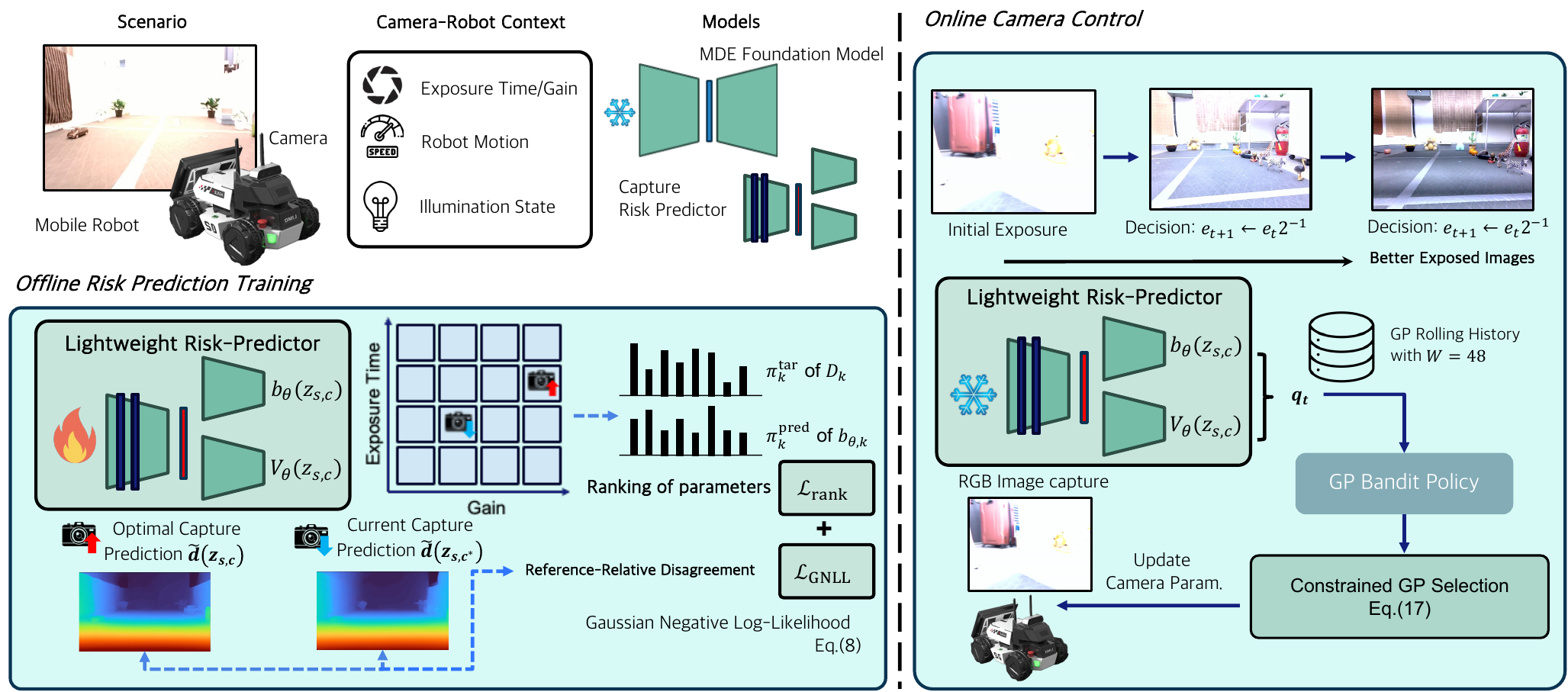}
    \vspace{-1mm}
    \caption{
    A lightweight FiLM~\cite{perez2018film}-conditioned network uses frozen MDE features with camera, motion, and illumination information to predict the mean and log variance of reference-relative prediction disagreement.
    The target is constructed using the best evaluated reference capture
    and optimized using Gaussian negative log-likelihood and ranking
    losses.
    }
    \vspace{-4mm}
    \label{fig:risk_prediction}
\end{figure*}

We first examine how camera settings affect monocular depth estimation (MDE) using a dataset collected by sweeping exposure and gain. Sec.~\ref{sec:experiment_data_collection} describes the capture protocol and scene configurations.

\subsection{Reference-Relative Camera Selection Objective}
Let \(s\) denote the scene state, including the robot viewpoint and let \(\mathbf{c}=(e,g)\) be a camera configuration with exposure $e$ and gain $g$, and $x(s,\mathbf c)$ the acquired image. A frozen depth model $f_\phi$ produces $\hat d(s,\mathbf c) = f_\phi(x(s,\mathbf c))$. For relative-depth models, let $\tilde d(s,\mathbf c)$ denote the prediction after scale-and-shift alignment to the corresponding ground-truth depth; for metric-depth
models, $\tilde d(s,\mathbf c) = \hat d(s,\mathbf c)$. We define the image-level absolute relative error as
\begin{equation}
\begin{aligned}
{\rm Abs_{rel}}(s,\mathbf{c})
&=
\frac{1}{|\Omega|}
\sum_{p \in \Omega}
\frac{
    \left|
        d_{p}(s)
        -
        \tilde d_{p}(s,\mathbf{c})
    \right|
}{
    d_{p}(s)
},
\end{aligned}
\label{eq:absrel}
\end{equation}
where \(\Omega\) is the set of valid pixels and \(d_{p}\) is the ground-truth depth.  These alignment and error conventions are used for
reference selection and evaluation.

For descriptive analysis, we model the variation in \({\rm Abs_{rel}}(s,\mathbf c)\) as the sum of scene-dependent, camera-dependent, and scene--camera interaction effects:
\begin{equation}
\begin{aligned}
{\rm Abs_{rel}}(s,\mathbf{c})
={}& {\rm Abs}_{\rm rel}^{0}
+ {\rm Abs}_{\rm rel}^{\mathrm{scene}}(s)
+ {\rm Abs}_{\rm rel}^{\mathrm{camera}}(\mathbf{c}) \\
&+ {\rm Abs}_{\rm rel}^{\mathrm{scene}\times\mathrm{camera}}(s,\mathbf{c})
+ \varepsilon,
\end{aligned}
\label{eq:error_decomposition}
\end{equation}
where \({\rm Abs}_{\rm rel}^{0}\) is the overall mean and \(\varepsilon\) represents unexplained residual variation. 
For the evaluated configuration set $\mathbf C$, define the
best evaluated setting $\mathbf c^*(s)$ (i.e., reference) and each setting's  excess error $\Delta {\rm Abs}_{\rm rel}(s,\mathbf c)$ as
\begin{equation}
    \mathbf{c}^{*}(s)
    =
    \underset{\mathbf{c} \in \mathbf{C}}{\arg\min}\;
    {\rm Abs}_{\rm rel}(s,\mathbf{c}),
\label{eq:optimal_camera}
\end{equation}
\begin{equation}
\begin{aligned}
\Delta {\rm Abs}_{\rm rel}(s,\mathbf{c})
={}&
{\rm Abs}_{\rm rel}(s,\mathbf{c})
-
{\rm Abs}_{\rm rel}\bigl(s,\mathbf{c}^{*}(s)\bigr). 
\end{aligned}
\label{eq:camera_degradation}
\end{equation}
For a fixed scene state, subtraction cancels the overall mean and additive scene term, while retaining camera effects, scene-camera interactions, and residual differences. Since the reference error is constant, excess error $\Delta {\rm Abs}_{\rm rel}(s,\mathbf{c})$ preserves their ground-truth error ranking. 
Our registered capture groups approximate this fixed-state comparison. Thus, Eq.~\eqref{eq:camera_degradation} defines a \textit{relative capture-quality criterion} without implying independence from scene type.

We use the same reference to construct \textbf{reference-relative disagreement (RRD)}, which measures the normalized difference between candidate and reference depth predictions. RRD is defined in Eq.~\eqref{eq:risk_target}. Ground-truth depth determines the optimal reference ($\mathbf c^*(s)$), while disagreement measures how the frozen model’s prediction changes across acquisitions. We investigate RRD as a supervision target later. 


 \subsection{Descriptive Variance Analysis}
%
In Table~\ref{tab:anova_target_comparison}, we analyze the variance of the camera-sweep responses using a two-way ANOVA with scene type and camera configuration as factors. To reduce dependence among adjacent frames, we aggregate frame-level measurements into lap-level responses after marginalizing illumination and motion, yielding 255 observations over four scene types (two training and two validation scenes) and 16 camera configurations. 

\begin{table}[t]
    \centering
    \caption{
    Descriptive two-way ANOVA of camera-sweep responses.
    $\Delta$\absrel denotes the ground-truth excess depth error, whereas \textbf{RRD} is the reference-relative disagreement.
    }
    \label{tab:anova_target_comparison}
    \resizebox{\columnwidth}{!}{
    \begin{tabular}{llccccc}
        \toprule
        Target & Effect & $df$ & $F$ & $p$-value & $\eta^2$ & Partial $\eta^2$ \\
        \midrule

        \multirow{3}{*}{\textbf{RRD}}
        & Scene
        & 3
        & 3.63
        & 0.0139
        & 0.71\%
        & 5.40\% \\

        & Camera
        & 15
        & 78.17
        & $<10^{-6}$
        & \textbf{76.27\%}
        & 85.99\% \\

        & Scene$\times$Camera
        & 45
        & 3.64
        & $<10^{-6}$
        & \textbf{10.65\%}
        & 46.16\% \\

        \midrule

        \multirow{3}{*}{$\Delta$\absrel}
        & Scene
        & 3
        & 5.30
        & 0.0016
        & 0.81\%
        & 7.69\% \\

        & Camera
        & 15
        & 92.29
        & $<10^{-6}$
        & \textbf{70.38\%}
        & 87.88\% \\

        & Scene$\times$Camera
        & 45
        & 8.35
        & $<10^{-6}$
        & \textbf{19.10\%}
        & 66.30\% \\ 
        \bottomrule
    \end{tabular}
    }
    \vspace{-2mm}
\end{table}

For $\Delta$\absrel, camera configuration accounts for 70.38\% of the reported variance, scene type for 0.81\%, and their interaction for 19.10\%. For RRD, the corresponding shares are 76.27\%, 0.71\%, and 10.65\%. Both responses exhibit substantial camera-associated variation together with scene-dependent camera effects. RRD has a larger camera-main-effect share and a smaller interaction share in this aggregated dataset.
These descriptive results motivate investigating \textit{RRD as a camera-selection target}. Its practical value depends on whether the learned score selects low-error acquisitions, which we evaluate through the selection experiments and supervision ablations in Sec. V.

\section{Method}
\label{sec:method}

\subsection{Overview}

Consider a mobile robot with a \textit{configurable} RGB camera and a \textit{frozen} MDE model \(f_\phi\). At online acquisition step \(t\), the robot captures an RGB image \(x_t\) using \(\mathbf c_t=(e_t,g_t)\in\mathbf C\). We collect the predictor inputs as 
\begin{equation}
z_t = (v_t,\mathbf c_t, \mathbf \xi_t), ~~ \mathbf \xi_t = (m_t, \ell_t), 
\end{equation} 
where \(m_t\) and \(\ell_t\) denote the externally supplied motion and illumination states, respectively, and \(v_t\) denotes the encoder feature representation extracted from \(x_t\) by the frozen MDE model.
The predictor scores the acquired image $x_t$, and a contextual GP uses completed feedback to select the next camera setting $\mathbf c_{t+1}$ from an admissible subset \(\mathbf C_t\subseteq\mathbf C\). Unacquired candidate images are not required. The predictor is trained offline (Sec. IV-B), while online adaptation updates the GP posterior (Sec. IV-C).

\subsection{Reference-Relative Feedback Prediction}
\label{sec:risk_prediction}

\noindent\textbf{Predictor and Supervision.} \qquad 
A lightweight FiLM-conditioned network (Fig. 2) combines frozen MDE encoder features with camera, motion, and illumination context.  It predicts a mean RRD \(b_\theta(z_t)\) and log-variance \(V_\theta(z_t)=\log\sigma_\theta^2(z_t)\) under the scalar observation model
\begin{equation}
D_t\mid z_t
\sim
\mathcal{N}
\left(
    b_{\theta}(z_t),
    \sigma_{\theta}^{2}(z_t)
\right),
\label{eq:risk_distribution}
\end{equation}
where $D_t$ denotes the reference-relative disagreement (RRD) associated with the capture.
Online, $D_t$ is unobserved and the network evaluates only its predicted mean and variance.

%
Offline, let $s$ index a registered training state and $\mathbf c^*(s)$ its lowest-error reference setting in its registered capture group from Eq. (3). Registration criteria are given in Sec. V-A.1. 
Define the predictor input as $z_{s,\mathbf c} = (v(s,\mathbf c),\mathbf c,\mathbf \xi_{s,\mathbf c})$, where $\mathbf \xi_{s,\mathbf c}$ is its recorded motion and illumination context. The candidate and reference images yield depth predictions \(\tilde d(z_{s,c})\doteq f_\phi(x(s,c))\), and \(\tilde d(z_{s,\mathbf c^*(s)})\)
For relative-depth models, we independently scale- and shift-align each prediction to its corresponding ground-truth depth, obtaining \(\tilde d(z_{s,\mathbf c})\) and \(\tilde d(z_{s,\mathbf c^*(s)})\); for metric-depth models, these equal the unaligned predictions. The RRD target $D(z_{s,\mathbf c})$ is
%
\begin{equation}
D(z_{s,\mathbf c})
=
\frac{1}{|\Omega|}
\sum_{p\in\Omega}
\frac{
    \left|
        \tilde d_p(z_{s,\mathbf c})-\tilde d_p(z_{s,\mathbf c^*(s)})
    \right|
}{
    \tilde d_p(z_{s,\mathbf c^*(s)})
},
\label{eq:risk_target}
\end{equation}
where $\Omega$ contains the valid candidate-reference comparison pixels. 
%
Ground truth and reference
captures are used offline for target construction and evaluation.




\noindent\textbf{Learning Objective.} \qquad 
We fit the scalar observation model 
using Gaussian negative log-likelihood, omitting the constant term: 
\begin{equation}
\mathcal{L}_{\mathrm{NLL}}
=
\frac{1}{2}
\left[
    e^{-V_{\theta}(z_{s,\mathbf c})}
    \left(
        D(z_{s,\mathbf c})
        -
        b_{\theta}(z_{s,\mathbf c})
    \right)^2
    +
    V_{\theta}(z_{s,\mathbf c})
\right].
\label{eq:risk_nll}
\end{equation}
We also use a listwise ranking loss to align candidate-score orderings. For a mini-set $\mathbf C_{mini}$ of candidate configurations at the same registered training state $s$, 
we define
\begin{equation}
\pi_\mathbf c^{\mathrm{tar}}
=
\frac{
    e^{D(z_{s,\mathbf c})/T}
}{
    \sum_{\mathbf c' \in \mathbf C_{mini}}
    e^{D(z_{s,\mathbf c'})/T}
},
~~~
\pi_\mathbf c^{\mathrm{pred}}
=
\frac{
    e^{b_\theta(z_{s,\mathbf c})/T}
}{
    \sum_{\mathbf c' \in \mathbf C_{mini}}
    e^{b_{\theta}(z_{s,\mathbf c'})/T}
},
\label{eq:ranking_distributions}
\end{equation}
and their cross-entropy:
\begin{equation}
\mathcal{L}_{\mathrm{rank}}
=
-
\sum_{\mathbf c \in \mathbf C_{mini}}
\pi_\mathbf c^{\mathrm{tar}}
\log \pi_\mathbf c^{\mathrm{pred}}.
\label{eq:ranking_loss}
\end{equation}
The positive logits assign larger training weights to captures with greater disagreement. Lower predicted RRD remains preferable for camera selection. The combined objective is
\begin{equation}
\mathcal{L}=\mathcal{L}_{\mathrm{NLL}}+\lambda_{\mathrm{rank}}\mathcal{L}_{\mathrm{rank}}
\end{equation}
where $|\mathbf C_{mini}|=8$,  \(\lambda_{\mathrm{rank}}=0.1\), and \(T=0.02\). Both the depth model and the trained predictor remain fixed during deployment. For a batch of \(B\) mini-sets, each of size $|\mathbf C_{mini}|$, \(\mathcal{L}_{\mathrm{NLL}}\) is averaged over the \(B|\mathbf C_{mini}|\) sample-level terms, whereas \(\mathcal{L}_{\mathrm{rank}}\) is averaged over the \(B\) groups.


\begin{table*}[t]
\centering
\caption{
Offline camera-selection performance.
Absolute regret is $R_{\mathrm{abs}}=\text{Abs}_{\text{rel}}(\hat d(z_{\mathbf c}))-\text{Abs}_{\text{rel}}(\hat d(z_{\mathbf c^{*}}))$, and relative regret is \(R_{\mathrm{rel}}=100R_{\mathrm{abs}}/\text{Abs}_{\text{rel}}(\hat d(z_{\mathbf c^{*}}))\). Best results are shown in bold.
}
\label{tab:offline_selection}
\resizebox{0.9\textwidth}{!}{
\begin{tabular}{lllcccccc}
\toprule
\multirow{2}{*}{Backbone}
& \multirow{2}{*}{Method}
& \multirow{2}{*}{Scene}
& \multicolumn{2}{c}{Corr.$(\Delta\text{Abs}_{\text{rel}},b_t)~\uparrow$}
& \multicolumn{2}{c}{Selection Regret$~\downarrow$}
& \multicolumn{2}{c}{Selection Accuracy$~\uparrow$} \\
\cmidrule(lr){4-5}
\cmidrule(lr){6-7}
\cmidrule(lr){8-9}

& & &
Pearson
& Spearman
& Absolute
& Relative (\%)
& \(R_{\mathrm{rel}}<10\%\)
& \(R_{\mathrm{rel}}<15\%\) \\
\midrule

\multirow{6}{*}{DAv2}
& \multirow{2}{*}{Ours}
& Training
& \textbf{0.88574}
& \textbf{0.83287}
& \textbf{0.003659}
& \textbf{5.985}
& \textbf{0.7624}
& \textbf{0.8581} \\

& & Validation
& \textbf{0.73107}
& \textbf{0.69427}
& \textbf{0.009819}
& \textbf{11.456}
& \textbf{0.6120}
& \textbf{0.7241} \\
\cmidrule(lr){2-9}

& \multirow{2}{*}{TTA~\cite{mi2022training}}
& Training
& 0.48893
& 0.54558
& 0.012822
& 19.0717
& 0.4822
& 0.5531 \\

& & Validation
& 0.28489
& 0.49525
& 0.028686
& 27.1782
& 0.4137
& 0.5258 \\
\cmidrule(lr){2-9}

& \multirow{2}{*}{Grad.~\cite{hornauer2022gradient}}
& Training
& 0.40268
& 0.27731
& 0.027805
& 42.3995
& 0.2907
& 0.3829 \\

& & Validation
& 0.29728
& 0.32117
& 0.047005
& 59.2078
& 0.2844
& 0.3448 \\
\midrule

\multirow{6}{*}{\shortstack{Metric-\\DAv2}}
& \multirow{2}{*}{Ours}
& Training
& \textbf{0.90723}
& \textbf{0.69222}
& \textbf{0.001514}
& \textbf{7.156}
& \textbf{0.7692}
& \textbf{0.8247} \\

& & Validation
& \textbf{0.95098}
& \textbf{0.64873}
& \textbf{0.003714}
& \textbf{12.990}
& \textbf{0.5752}
& \textbf{0.7079} \\
\cmidrule(lr){2-9}

& \multirow{2}{*}{TTA~\cite{mi2022training}}
& Training
& 0.77269
& 0.42144
& 0.003751
& 18.195
& 0.5256
& 0.6495 \\

& & Validation
& 0.31304
& 0.34815
& 0.048776
& 189.759
& 0.4778
& 0.5221 \\
\cmidrule(lr){2-9}

& \multirow{2}{*}{Grad.~\cite{hornauer2022gradient}}
& Training
& 0.10023
& -0.00551
& 0.018427
& 83.104476
& 0.2094
& 0.3076 \\

& & Validation
& -0.14095
& -0.13952
& 0.041483
& 167.54618
& 0.1504
& 0.2654 \\
\bottomrule
\end{tabular}
}
\vspace{-5mm}
\end{table*}

\subsection{Online Contextual GP Camera Control}
\label{sec:online_camera_control}

We formulate exposure and gain selection as a finite-action contextual GP bandit. 
Let \(\mathcal H_t\) contain the latest \(W\) valid completed observations available when GP conditioning begins. 

\noindent\textbf{Acquisition Feedback.} \qquad
For each acquired image, the predictor provides the risk-sensitive score $q_t$: 
\begin{equation}
q_t
=
b_\theta(z_t)+\kappa\sigma_\theta(z_t),
\qquad \kappa\geq0.
\label{eq:gp_risk_observation}
\end{equation}
The second term penalizes dispersion in predicted RRD using a weight factor $\kappa$. The resulting score is an acquisition cost, not a calibrated probability of perception failure. 
Each record $(\mathbf u_t, q_t)$ uses the GP input \(\mathbf{u}_t=(\mathbf{c}_t,\boldsymbol{\xi}_t,\tau_t)\), where \(\tau_t\) is the capture timestamp.

\noindent\textbf{GP Model.} \qquad
Let \(\bar q_t\) and \(\sigma_{q,t}\) denote the mean and standard deviation of the scores in \(\mathcal H_t\). For each retained record $(\mathbf u_i, q_i) \in \mathcal H_t $, we standardize the score and model it as 
\begin{equation}
\begin{aligned}
\gamma_t
&=\max\{\sigma_{q,t},\gamma_{\min}\},
&\widetilde q_{i|t}
&=\frac{q_i-\bar q_t}{\gamma_t},\\
\widetilde h_t
&\sim\mathrm{GP}(0,k),
&\widetilde q_{i|t}
&=\widetilde h_t(\mathbf{u}_i)+\epsilon_i,\quad
\epsilon_i\sim\mathcal{N}(0,\sigma^2_n).
\end{aligned}
\label{eq:online_gp_model}
\end{equation}
Here, $\gamma_{\min}>0$ prevents degenerate standardization and  \(\sigma^2_n\) is residual observation-noise variance on the standardized scale. The index $i | t$ indicates normalization of observation $i$ using the history at decision $t$. 


%
Following GP-UCB~\cite{srinivas2010gpucb} and contextual GP bandit optimization~\cite{krause2011contextualgp}, we employ the product kernel
\begin{equation}
k(\mathbf{u}_i,\mathbf{u}_j)
=
k_c(\mathbf{c}_i,\mathbf{c}_j)
k_m(m_i,m_j)
k_l(\ell_i,\ell_j)
k_\tau(\tau_i,\tau_j).
\label{eq:contextual_gp_kernel}
\end{equation}
The camera kernel $k_c$ is an anisotropic squared-exponential over normalized indices of exposure and gain in $\mathbf C$ with length scales $\lambda_e,\lambda_g$. $k_m$ and $k_l$ are categorical kernels for motion and illumination with unit covariance of identical states and $\rho_m,\rho_l$ otherwise. The temporal kernel $k_\tau(\tau_i,\tau_j) =\exp(-|\tau_i-\tau_j|/\lambda_\tau)$ decreases covariance with temporal separation. 

For candidate $\mathbf c$, let \(\tilde\mu_t^{\mathrm{GP}}(\mathbf c)\) and \(\tilde\sigma_t^{\mathrm{GP}}(\mathbf c)\) denote posterior moments at the context and timestamp supplied to the current decision. We restore the original score scale using 
\begin{equation}
\mu^{\mathrm{GP}}_t(\mathbf c) = \bar q_t + \gamma_t \tilde \mu^{\mathrm{GP}}_t(\mathbf c), ~~~ \sigma^{\mathrm{GP}}_t(\mathbf c) = \gamma_t \tilde \sigma^{\mathrm{GP}}_t(\mathbf c). 
\end{equation}


\noindent\textbf{Constrained Selection.} \qquad
A motion-dependent exposure envelope and a photometric guard restrict the candidate set to $\mathbf C_t$. The guard limits exposure-value changes and screens projected image statistics. For candidate $\mathbf c$, it approximates luminance by 
%
\begin{equation}
\Delta EV_t(\mathbf c) = {\rm{log}}_2 \frac{eg}{e_t g_t}, ~~~ \hat Y_{t,p}(\mathbf c) = Y_{t,p} 2^{\Delta EV_t(\mathbf c)}
\end{equation}
%
with corresponding projected RGB values used for saturation screening. 
Exposure caps and photometric thresholds are specified in Sec.~\ref{sec:experiment_implement_details}. 
When \(\mathbf C_t\) is nonempty and GP conditioning succeeds, the controller selects
\begin{equation}
\begin{aligned}
\mathbf{c}_{t+1}
=\arg\min_{\mathbf{c}\in\mathbf{C}_t}
\Bigl[
&\mu_t^{\mathrm{GP}}(\mathbf{c})
-\beta_{\mathrm{GP}}\sigma_t^{\mathrm{GP}}(\mathbf{c}) 
+\lambda_{\mathrm{switch}}
\mathbf{1}\{\mathbf{c}\neq\mathbf{c}_t\}
\Bigr].
\end{aligned}
\label{eq:gp_selection}
\end{equation}
Here, the GP uncertainty term encourages exploration, whereas the switching penalty discourages unnecessary configuration changes. This uncertainty term serves a different role from $\sigma_\theta$ in Eq. (12), which increases the observed acquisition cost. 

\noindent\textbf{Asynchronous Update.} \qquad 
GP conditioning overlaps predictor inference on the current capture and uses only completed observations in \(\mathcal H_t\). After selection, the controller issues the next setting to the camera. When predictor inference finishes, a valid score is appended with its own capture configuration, context, and timestamp, retaining at most \(W\) observations. Online adaptation therefore updates the GP posterior while keeping the neural models fixed. 
Controller hyperparameters, initialization, and fallback behavior are described in Sec.~\ref{sec:experiment_implement_details}.
\section{Experiments}
\label{sec:experiments}

\subsection{Experimental Setup}

\subsubsection{Data Collection}
\label{sec:experiment_data_collection}
We collected RGB-D sequences using an AgileX Robotics LIMO Pro equipped with an Orbbec 336 camera and an NVIDIA Jetson Orin Nano. The robot repeatedly traversed a fixed trajectory in both directions at a maximum linear speed of \(1.0\,\mathrm{m/s}\) and angular speed \(1.0\,\mathrm{rad/s}\). We evaluated the 16 exposure--gain configurations in the discrete grid \(\mathbf{C}=(\{4,8,16,32\}\,\mathrm{ms})\times\{16,32,64,128\}\). 
The dataset contains approximately \(51\mathrm{K}\) frames, partitioned at the scene-configuration level into three training and three validation configurations, with all runs and traversal laps from each configuration assigned exclusively to one split. Each training configuration shares the structural layout of its validation counterpart but differs in object identity and arrangement. Validation Scene~1 uses an enclosed layout formed by temporary walls, Validation Scene~2 removes these walls, and Validation Scene~3 retains the open layout with additional textured wall coverings. Illumination was categorized as bright (\(\approx70\,\mathrm{lux}\)), normal
(\(\approx10\,\mathrm{lux}\)), or dark (\(<5\,\mathrm{lux}\)). Based on linear speed \(|v|\), translational motion was categorized as stop (\(<0.05\,\mathrm{m/s}\)), slow (\(0.05\)--\(0.25\,\mathrm{m/s}\)), or fast
(\(0.25\)--\(1.0\,\mathrm{m/s}\)). When translation was negligible, rotational motion was categorized as rotate or spin using an angular speed threshold of \(0.5\,\mathrm{rad/s}\). Motion states are derived from the wheel encoders of the LIMO Pro, whereas illumination states are manually assigned.

Within each scene, observations depicting approximately the same viewpoint under different camera configurations were registered using ground-truth depth, requiring at least 95\% spatial overlap, an ECC score above 0.9, and maximum 32px translation error; failed registrations were discarded. The configuration with the lowest measured \absrel was treated as the best evaluated reference, and the remaining configurations were treated as candidates.

\subsubsection{Implementation Details}
\label{sec:experiment_implement_details}

We use Depth Anything v2 Small at \(640\times 480\) and extract
encoder features for a risk predictor containing 165,910 parameters. The predictor is trained using AdamW with learning rate $5\times 10^{-4}$ and batch size $B=4$. The controller uses \(\kappa=1.645\), \(\beta_{\mathrm{GP}}=0.5\), \(\lambda_{\mathrm{switch}}=0.005\), \(W=48\), \(\sigma^2_n=0.1\), \(\lambda_{\tau}=5.0, \lambda_e=\lambda_g=0.35\), \(\rho_m=\rho_l=0.2\), \(\gamma_{\text{min}}=0.01\). 
When history is available, GP conditioning succeeds, and \(\mathbf C_t\) is nonempty, the controller uses Eq. (17). If the GP history is empty or  conditioning fails, it retains the current admissible configuration; otherwise, it selects the admissible configuration closest to the configured default. 
For the admissibility filters, the motion filter caps exposure time at \(16\,\mathrm{ms}\) for fast and rotate, and \(8\,\mathrm{ms}\) for spin. The photometric guard detects overexposure when the saturated-pixel ratio reaches \(0.89\) with a projected mean luminance $\hat Y_{t+1}$ of at least \(243\), and detects hard underexposure when the shadow-pixel ratio reaches \(0.50\) with a mean luminance of at most \(45\).


\begin{table*}[t]
\centering
\caption{
Online MDE performance across validation scenes and illumination conditions using affine-invariant Depth Anything v2. Values are the mean \(\pm\) standard deviation of trial-level
aligned \absrel and \(\delta_{1.25}\) over five repetitions per condition. Bold indicates the best mean in each condition. Standard deviations are rounded to three decimal places.
}
\vspace{-2mm}
\label{tab:online_mde_validation}
\resizebox{\textwidth}{!}{%
\begin{tabular}{llcccccc}
\toprule
\multirow{2}{*}{Illumination}
& \multirow{2}{*}{Method}
& \multicolumn{2}{c}{Validation Scene 1}
& \multicolumn{2}{c}{Validation Scene 2}
& \multicolumn{2}{c}{Validation Scene 3} \\
\cmidrule(lr){3-4}
\cmidrule(lr){5-6}
\cmidrule(lr){7-8}
&
& \(\text{Abs}_{\text{rel}}\downarrow\)
& \(\delta_{1.25}\uparrow\)
& \(\text{Abs}_{\text{rel}}\downarrow\)
& \(\delta_{1.25}\uparrow\)
& \(\text{Abs}_{\text{rel}}\downarrow\)
& \(\delta_{1.25}\uparrow\) \\
\midrule

\multirow{4}{*}{Bright}
& AE
& \(0.07171{\pm}0.003\)
& \(0.92343{\pm}0.006\)
& \(0.07908{\pm}0.001\)
& \(0.92179{\pm}0.002\)
& \(0.07103{\pm}0.002\)
& \(0.93458{\pm}0.006\) \\

& Shin et al.~\cite{shin2019camera}
& \(0.09171{\pm}0.024\)
& \(0.88528{\pm}0.044\)
& \(0.14344{\pm}0.058\)
& \(0.85261{\pm}0.057\)
& \(0.10813{\pm}0.024\)
& \(0.88608{\pm}0.034\) \\

& Zhang et al.~\cite{zhang2024efficient}
& \(0.05395{\pm}0.002\)
& \(0.95912{\pm}0.003\)
& \(0.08259{\pm}0.002\)
& \(0.91938{\pm}0.002\)
& \(0.07849{\pm}0.002\)
& \(0.92341{\pm}0.005\) \\

& Ours
& \(\mathbf{0.04776{\pm}0.001}\)
& \(\mathbf{0.96271{\pm}0.004}\)
& \(\mathbf{0.06706{\pm}0.001}\)
& \(\mathbf{0.94124{\pm}0.002}\)
& \(\mathbf{0.06755{\pm}0.001}\)
& \(\mathbf{0.94017{\pm}0.001}\) \\
\midrule

\multirow{4}{*}{Normal}
& AE
& \(0.05672{\pm}0.001\)
& \(0.94998{\pm}0.004\)
& \(0.08280{\pm}0.004\)
& \(0.92703{\pm}0.004\)
& \(0.08074{\pm}0.003\)
& \(0.92618{\pm}0.003\) \\

& Shin et al.~\cite{shin2019camera}
& \(0.04416{\pm}0.002\)
& \(0.96934{\pm}0.003\)
& \(0.09184{\pm}0.004\)
& \(0.92293{\pm}0.004\)
& \(0.09702{\pm}0.007\)
& \(0.91548{\pm}0.010\) \\

& Zhang et al.~\cite{zhang2024efficient}
& \(\mathbf{0.04410{\pm}0.003}\)
& \(\mathbf{0.97218{\pm}0.006}\)
& \(0.08262{\pm}0.005\)
& \(\mathbf{0.93373{\pm}0.004}\)
& \(0.09178{\pm}0.007\)
& \(0.93029{\pm}0.007\) \\

& Ours
& \(0.04528{\pm}0.001\)
& \(0.97084{\pm}0.003\)
& \(\mathbf{0.07997{\pm}0.003}\)
& \(0.93027{\pm}0.004\)
& \(\mathbf{0.07878{\pm}0.006}\)
& \(\mathbf{0.93430{\pm}0.008}\) \\
\midrule

\multirow{4}{*}{Dark}
& AE
& \(0.05438{\pm}0.010\)
& \(0.96238{\pm}0.000\)
& \(0.07582{\pm}0.007\)
& \(0.92429{\pm}0.012\)
& \(0.06633{\pm}0.003\)
& \(0.94086{\pm}0.004\) \\

& Shin et al.~\cite{shin2019camera}
& \(0.04293{\pm}0.001\)
& \(0.97307{\pm}0.002\)
& \(0.07302{\pm}0.004\)
& \(0.92958{\pm}0.008\)
& \(0.07398{\pm}0.003\)
& \(0.92893{\pm}0.008\) \\

& Zhang et al.~\cite{zhang2024efficient}
& \(0.04334{\pm}0.003\)
& \(0.97381{\pm}0.006\)
& \(0.06049{\pm}0.002\)
& \(0.95633{\pm}0.003\)
& \(0.06873{\pm}0.004\)
& \(0.94913{\pm}0.005\) \\

& Ours
& \(\mathbf{0.04082{\pm}0.000}\)
& \(\mathbf{0.97652{\pm}0.001}\)
& \(\mathbf{0.05815{\pm}0.002}\)
& \(\mathbf{0.95806{\pm}0.002}\)
& \(\mathbf{0.06022{\pm}0.001}\)
& \(\mathbf{0.95916{\pm}0.002}\) \\
\bottomrule
\end{tabular}%
}
\vspace{-3mm}
\end{table*}

\subsubsection{Evaluation Metrics}

During training, candidate mini-sets with \(|\mathbf C_{mini}|=8\) were used for the
ranking loss, whereas offline evaluation used groups of \(|\mathbf C_{mini}|=10\).
Let \({\rm Abs}^{i}_{\rm rel}\) denote the measured \absrel of observation
\(i\), and let \(\hat{b}_i\) denote its predicted score, where lower
values are preferred. The predictor and oracle selections are defined as
\[
\hat{i}_g=\arg\min_{i\in\mathbf{C}_{mini}}\hat{b}_i,
\qquad
i_g^*=\arg\min_{i\in\mathbf{C}_{mini}}{\rm Abs}^{i}_{\rm rel}.
\]

For each observation \(i\in\mathbf{C}_{mini}\), the correlation target is
the excess error $\Delta {\rm Abs}^{i}_{\rm rel} = {\rm Abs}^{i}_{\rm rel}-{\rm Abs}^{i_g^*}_{\rm rel}$, corresponding to \absrel in Table~\ref{tab:offline_selection}. Pearson and Spearman correlations between \(\hat{b}_i=b_\theta(z_i)\) and \(\Delta {\rm Abs}^i_{\rm rel}\) are computed globally over all samples in each evaluation partition. Normalized groupwise selection regret and tolerance accuracy are defined as 
\begin{equation}
\begin{aligned}
\widetilde{R}_g
&=
\frac{\Delta {\rm Abs}^{\hat i_g}_{\rm rel}}{{\rm Abs}^{i_g^*}_{\rm rel}},\quad
\operatorname{Acc}_{\mathrm{group}}(\tau)
&=
\frac{1}{N_G}
\sum_{g=1}^{N_G}
\mathbb{I}\!\left[\widetilde{R}_g<\tau\right],
\end{aligned}
\label{eq:selection_metrics}
\end{equation}
where \(N_G\) is the number of evaluation groups. We report mean
groupwise regret and accuracy at
\(\tau\in\{0.10,0.15\}\). For affine-invariant Depth Anything v2, each predicted depth map is scale- and shift-aligned in inverse-depth space to the corresponding ground-truth depth before evaluation. No alignment is applied to Metric Depth Anything v2. Online performance is evaluated using the conventional \absrel and \(\delta_{1.25}\) metrics under the same alignment conventions. Ground-truth depth is used only for training target construction and evaluation, and is never provided to the online controller.


\subsection{Offline Selection Performance via Risk Prediction}

We compare our learned risk predictor with two training-free uncertainty baselines for monocular depth estimation: 
test-time augmentation (TTA)~\cite{mi2022training} and gradient-based uncertainty~\cite{hornauer2022gradient}. As reported in Table~\ref{tab:offline_selection}, our method consistently achieves the highest correlations,  lowest selection regret, and highest selection accuracy for both the affine-invariant  and metric variants of Depth Anything v2~\cite{yang2024depth}. Across all backbone and scene combinations, our method reduces absolute selection regret by \(59.6\%\)--\(92.4\%\) relative to the uncertainty baselines. 
This advantage remains substantial in validation scenes. 
For example, with Metric-DAv2, our method achieves a Pearson correlation of \(0.951\), compared with \(0.313\) for TTA and \(-0.141\) for gradient-based uncertainty. 
These results indicate that the learned risk provides a more reliable ranking of camera configurations under the evaluated scene shifts than the training-free baselines.

\subsection{Online Selection Performance}
We compare our method with the camera's built-in auto-exposure (AE) and the online exposure controllers of Shin et al.~\cite{shin2019camera} and Zhang et al.~\cite{zhang2024efficient}. For Zhang et al.~\cite{zhang2024efficient}, we used the publicly released DRL-Feat checkpoint, which achieved the best reported performance among their reward variants. For all baseline methods, we retained the original action spaces and hyperparameter settings. Table~\ref{tab:online_mde_validation} reports MDE performance across three validation scenes and three illumination conditions. Each condition was repeated five times, and the results are reported as the mean and std.

\begin{figure}[!t]
    \centering
    \includegraphics[width=\columnwidth]{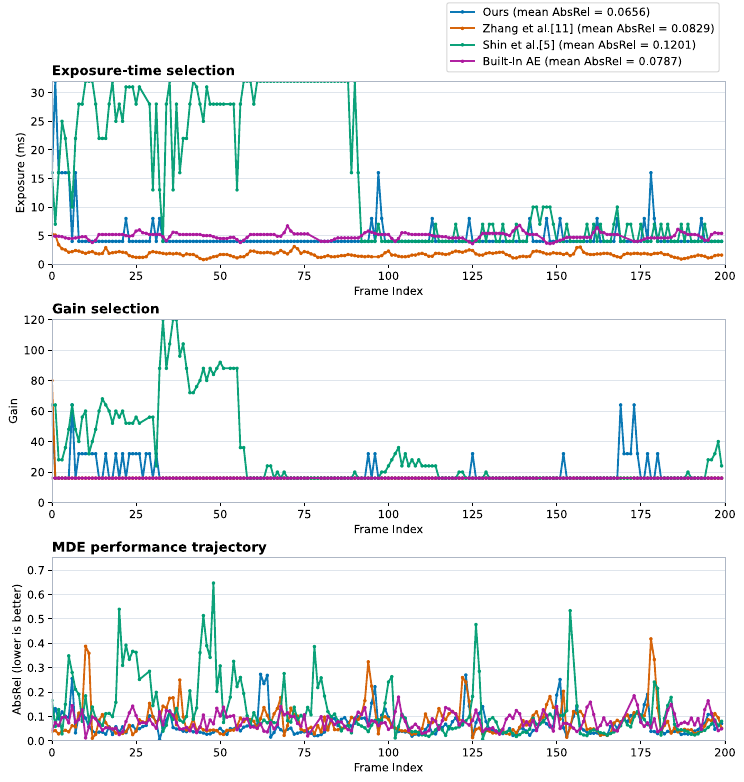}
    \vspace{-7mm}
    \caption{
    Online trajectories of exposure time, gain, and \absrel under validation scene 2 in bright illumination. Our policy achieves a lower mean \absrel ($0.0656$) and avoids frequent error spikes associated with wrong camera configurations. 
    }
    \label{fig:online_trajectory}
    \vspace{-1.5mm}
\end{figure}

\ours{} achieves the lowest mean AbsRel in eight of nine conditions and the highest mean $\delta_{1.25}$ in seven. Averaging the nine condition means with equal weight yields AbsRel values of 0.06062 for \ours{}, 0.07096 for built-in AE, and 0.06734 for Zhang et al.~\cite{zhang2024efficient}, corresponding to reductions of 14.6\% and 10.0\%, respectively. Under normal illumination, Zhang et al.~\cite{zhang2024efficient} performs best in Validation Scene 1 and achieves the highest $\delta_{1.25}$ in Scene 2. Fig.~\ref{fig:online_trajectory} illustrates the configuration changes and depth errors observed for the evaluated controllers over a representative sequence. 

\begin{figure*}[t]
    \centering
    \includegraphics[width=\textwidth]{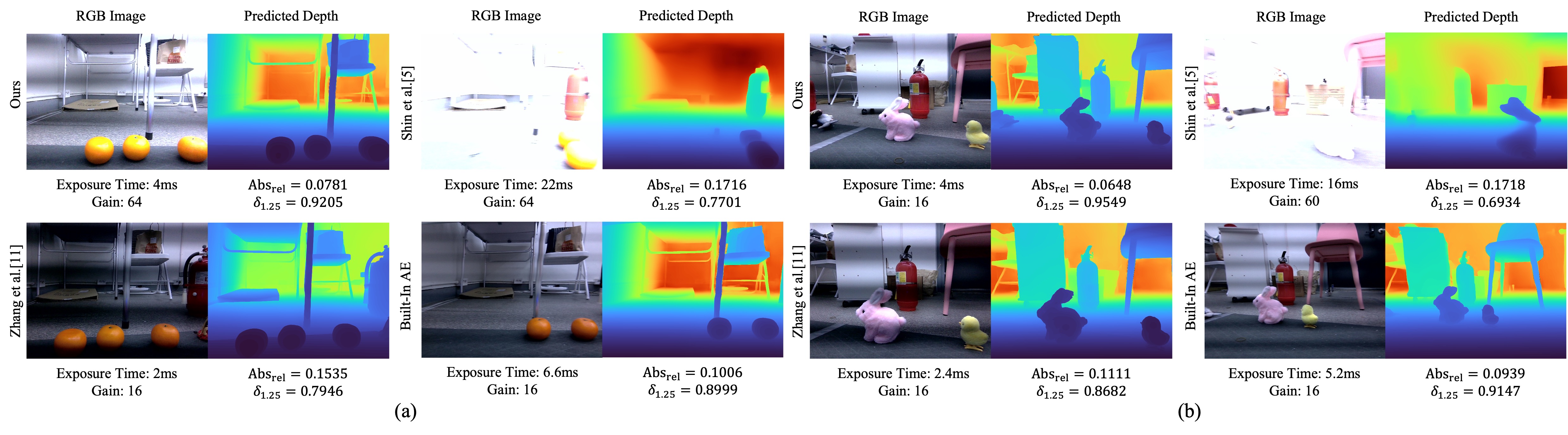}
    \vspace{-7mm}
    \caption{
    Qualitative comparison of online camera selections in two different views.
    RGB images and MDE predictions are shown for Ours,
    Zhang et al.~\cite{zhang2024efficient}, Shin et al.~\cite{shin2019camera},
    and built-in AE.
    }
    \label{fig:qualitative_selection}
    \vspace{-4mm}
\end{figure*}

Qualitatively, the error trajectories in Fig.~\ref{fig:online_trajectory} further illustrate these differences. 
Camera configurations that do not account for the time-varying scene--camera interaction, produce pronounced MDE error spikes. 
Shin et al.~\cite{shin2019camera} requires a relatively long search to approach a single optimum, during which long exposures can cause severe overexposure and degrade MDE performance. 
Conversely, Zhang et al.~\cite{zhang2024efficient} frequently selects low exposure and gain settings in darker views under the bright condition, resulting in underexposed images. 

At the image level, Fig.~\ref{fig:qualitative_selection} compares captures from each method at a comparable viewpoint during \textit{spin} motion under bright illumination. Under flickering illumination, Zhang et al.~\cite{zhang2024efficient} selects low exposure and gain, yielding underexposure and a compressed depth range in distant regions. Shin et al.~\cite{shin2019camera} selects a long exposure, and overexposure with motion blur erases the table and chair, leaving only a smooth depth gradient. Built-in AE preserves the overall layout, but its longer exposure blurs thin structures such as the table leg. Our controller instead pairs a short exposure with appropriate gain to preserve brightness under flicker, retaining object boundaries in the predicted depth. 

Taken together, these results show that the proposed \textit{model-in-the-loop} framework selects camera configurations that are better aligned with MDE performance across varying scenes and illumination conditions. 
By explicitly modeling reference-relative disagreement caused by camera misconfiguration, \ours{} provides a more informative control signal than image-quality-based alternatives. The consistent quantitative and qualitative improvements also support our core assumption that reference-relative disagreement can serve as a practical signal for adaptive camera control.

 \subsection{Component Analysis}
We analyze the components of \ours: the supervision of the risk predictor, the admissibility constraints of the controller, and the cost of running both on the robot.

\noindent\textbf{Predictor Supervision.} \qquad 
We first ablate the two design choices of the risk predictor: we remove the cross-entropy ranking loss and replace the RRD target with direct ground-truth (GT)-based $\Delta$\absrel supervision. As shown in Table~\ref{tab:selection_ablation}, the GT-supervised variant achieves lower training regret, whereas RRD supervision yields the lowest validation regret. RRD reduces validation absolute regret from 0.013029 to 0.009819, a 24.6\% reduction relative to the GT-supervised variant. Removing the ranking loss increases validation absolute regret to 0.0146962; the full objective therefore reduces it by 33.2\%. These results support the utility of reference-relative supervision and ranking alignment in the evaluated training, validation scene configurations.


\begin{table}[t]
\centering
\caption{
Ablation of supervision and ranking loss for Depth Anything v2 (affine-invariant). GT excess-error supervision replaces Eq.~\eqref{eq:risk_target} with Eq.~\eqref{eq:camera_degradation}
and retains the ranking loss Eq.~\eqref{eq:ranking_loss}. The best result for each scene split is shown in bold.
}
\vspace{-2mm}
\label{tab:selection_ablation}
\setlength{\tabcolsep}{2.5pt}
\renewcommand{\arraystretch}{1.08}

\resizebox{\columnwidth}{!}{%
\begin{tabular}{llcccc}
\toprule
\multicolumn{2}{c}{Metric}
& \multicolumn{2}{c}{Selection Regret$~\downarrow$}
& \multicolumn{2}{c}{Selection Accuracy$~\uparrow$} \\
\cmidrule(lr){1-2}
\cmidrule(lr){3-4}
\cmidrule(lr){5-6}

Method
& Scene
& Absolute
& Relative (\%)
& \(R<10\%\)
& \(R<15\%\) \\
\midrule

\multirow{2}{*}{Ours}
& Training
& 0.003659
& 5.985
& 0.7624
& 0.8581 \\
& Validation
& \textbf{0.009819}
& \textbf{11.456}
& \textbf{0.6120}
& \textbf{0.7241} \\
\midrule

\multirow{2}{*}{\shortstack{Ours w/o\\ ranking loss}}
& Training
& 0.0064161
& 9.633
& 0.6205
& 0.7304 \\
& Validation
& 0.0146962
& 16.788
& 0.4827
& 0.5862 \\
\midrule

\multirow{2}{*}{\shortstack{Ours w/\\GT excess error}}
& Training
& \textbf{0.0025568}
& \textbf{4.226}
& \textbf{0.8297}
& \textbf{0.9219} \\
& Validation
& 0.0130290
& 14.758
& 0.5431
& 0.6637 \\
\bottomrule
\end{tabular}%
}
\vspace{-2mm}
\end{table}

\noindent\textbf{Admissibility Constraints.} \qquad 
Next, we isolate the contributions of the motion and photometric admissibility filters by removing each filter from the controller and evaluating the resulting variants in Validation Scene~1 under dark illumination. As shown in Table~\ref{tab:safety_filter_ablation}, enabling either filter improves both \absrel and \(\delta_{1.25}\), while their combination achieves the best performance. Compared with the variant without either filter, the complete configuration reduces \absrel by 4.69\%. It also achieves a 2.0\%--2.5\% reduction relative to the variants using only one filter. These results indicate that the two filters play complementary roles in restricting unfavorable exploratory actions. Although the risk-guided GP policy maintains reasonable performance without the filters, the admissibility constraints remain beneficial because GP exploration can otherwise select camera configurations that temporarily degrade perception quality.

\begin{table}[t]
\centering
\caption{
Ablation of the motion and photometric admissibility filters in Validation Scene 1 under dark illumination. Values are reported as mean \(\pm\) std over five repeated trials.
The best results are shown in bold.
}
\vspace{-2mm}
\label{tab:safety_filter_ablation}
\setlength{\tabcolsep}{8pt}
\renewcommand{\arraystretch}{1.08}

\resizebox{\columnwidth}{!}{%
\begin{tabular}{cccc}
\toprule
\multicolumn{2}{c}{Method}
& \multicolumn{2}{c}{Metric} \\
\cmidrule(lr){1-2}
\cmidrule(lr){3-4}

Motion
& Photometric
& \(\operatorname{AbsRel}\downarrow\)
& \(\delta_{1.25}\uparrow\) \\
\midrule

&
& \(0.04283{\pm}0.000\)
& \(0.97229{\pm}0.001\) \\

& \(\checkmark\)
& \(0.04166{\pm}0.000\)
& \(0.97559{\pm}0.001\) \\

\(\checkmark\) &
& \(0.04187{\pm}0.001\)
& \(0.97341{\pm}0.002\) \\

\(\checkmark\)
& \(\checkmark\)
& \(\mathbf{0.04082{\pm}0.000}\)
& \(\mathbf{0.97652{\pm}0.001}\) \\
\bottomrule
\end{tabular}%
}
\vspace{-1mm}
\end{table}

\begin{table}[t]
\centering
\caption{Serialized per-frame latency (ms). Total latency is the sum of the listed mean component latencies, and FPS is its reciprocal. Camera-switching latency is averaged over all frames.}
\vspace{-2mm}
\footnotesize
\setlength{\tabcolsep}{3pt}
\renewcommand{\arraystretch}{1.15}
\resizebox{\columnwidth}{!}{%
\begin{tabular}{lccccc}
\hline
Method & MDE & Image Processing & Control & Total & FPS \\
 & (ms) & (ms) & (ms) & (ms) & (Hz) \\
\hline
Built-in AE
& \multirow{4}{*}{205.82} & <1 & <1 & 205.82 & 4.858 \\

Shin et al.~[5]
& & 34.32 & 109.07 & 349.21 & 2.863 \\

Zhang et al.~[11]
& & -- & 116.57 & 322.39 & 3.101 \\

Ours
& & 26.32 & 66.13 & 298.27 & 3.352 \\
\hline
\end{tabular}%
}
\vspace{-2mm}
\label{tab:runtime_fps}
\end{table}

\noindent\textbf{Runtime.} \qquad 
Finally, to quantify the computational cost of both components on the robot, we measure serialized end-to-end latency with each camera controller and Depth Anything V2 implemented in PyTorch. As shown in Table~\ref{tab:runtime_fps}, ADEPT achieves \(298.27\pm11.15\,\mathrm{ms}\), lower than Shin et al.~\cite{shin2019camera} and Zhang et al.~\cite{zhang2024efficient}, but higher than built-in AE. MDE inference dominates the latency across all methods at \(205.82\,\mathrm{ms}\). The \(66.13\,\mathrm{ms}\) control latency of ADEPT comprises \(59.48\,\mathrm{ms}\) for camera switching and \(6.65\,\mathrm{ms}\) for decision computation. By reusing historical risk observations, filtering inadmissible configurations, and penalizing configuration changes, ADEPT helps avoid unnecessary transitions while retaining a decision cost comparable to Zhang et al.~(\(7.03\,\mathrm{ms}\)). Shin et al. incurs \(34.32\,\mathrm{ms}\) for image-quality computation and \(98.51\,\mathrm{ms}\) for camera switching, resulting in the lowest throughput among the evaluated controllers.
\section{CONCLUSION}
\label{sec:conclusion}

This paper presented a model-in-the-loop framework for adaptive
exposure and gain control for monocular depth estimation. We defined
excess ground-truth \absrel relative to the best
evaluated camera setting as the desired acquisition objective, while
training the predictor using reference-relative disagreement 
as a practical surrogate. A descriptive analysis of the
collected camera sweep shows that camera settings and their
interactions with scene type account for most of the observed
variation. In validation scenes, the proposed surrogate reduces
selection regret relative to direct ground-truth error supervision,
supporting its empirical transfer without assuming mathematical
equivalence to the desired objective. Integrated with a contextual GP
controller and admissibility constraints, the proposed feedback
improves mean online MDE performance over auto-exposure and the
evaluated prior controllers. These results support the practical utility
of reference-relative model feedback under the evaluated scene,
motion, and illumination conditions.

\bibliographystyle{IEEEtran}
\bibliography{references}

\end{document}